\RequirePackage{fix-cm}
\documentclass[twocolumn]{svjour3}          

\smartqed  
\usepackage{amsmath}
\usepackage{amssymb}
\usepackage{graphicx}
\usepackage{float}        
\usepackage{balance}      
\usepackage{multirow}     
\usepackage{hyperref}     
\usepackage{orcidlink}    
\usepackage{comment}      
\usepackage{array}
\usepackage{booktabs}     

\usepackage[table, dvipsnames]{xcolor}  
\definecolor{lgray}{gray}{0.9}          
\usepackage{soul}\sethlcolor{lgray}     
\usepackage{enumitem}                   
\newcolumntype{H}{>{\setbox0=\hbox\bgroup}c<{\egroup}@{}}  
\usepackage[normalem]{ulem}             

\usepackage{tikz}
\definecolor{tgblueF}{HTML}{E8F0FA}\definecolor{tgblueE}{HTML}{3A6EA5}
\definecolor{tggreenF}{HTML}{EEF2E3}\definecolor{tggreenE}{HTML}{5B7F2B}
\definecolor{tgtealF}{HTML}{E3F3F0}\definecolor{tgtealE}{HTML}{2A8C7E}
\definecolor{tgorangeF}{HTML}{FFF0DC}\definecolor{tgorangeE}{HTML}{C07F20}
\newcommand{\tg}[2]{\tikz[baseline=(t.base)]\node[draw=tg#2E, fill=tg#2F, rounded corners=2pt, inner sep=1.5pt, line width=0.4pt] (t) {#1};}  
\newcommand{\loc}[1]{\hl{#1}}
\newcommand{\sem}[1]{\uline{#1}}
\newcommand{\locsem}[1]{{\setlength{\fboxsep}{0pt}\colorbox{lgray}{\strut\uline{#1}}}}

\usepackage{microtype}    
\journalname{The VLDB Journal}
\let\makeheadbox\relax  

\begin{document}
\title{LimoBench: How Street Smart Are Large Language Models?}






\author{
Gautam Choudhary
    \and Jack Peng
    \and Udesh Kumarasinghe
    \and Rachel Rafik
    \and Yousef Hammad
    \and Ahmed Samy
    \and Walid G. Aref~\orcidlink{0000-0001-8169-7775}
}

\institute{
Gautam Choudhary~\orcidlink{0000-0002-0261-7851} \at Purdue University, West Lafayette, Indiana, USA \\ \email{gchoudha@purdue.edu}
  \and Jack Peng \at Purdue University, West Lafayette, Indiana, USA \\ \email{peng315@purdue.edu}
  \and Udesh Kumarasinghe~\orcidlink{0000-0002-3836-7495} \at Purdue University, West Lafayette, Indiana, USA \\ \email{ukumaras@purdue.edu}
  \and Rachel Rafik \at Purdue University, West Lafayette, Indiana, USA \\ \email{rrafik@purdue.edu}
  \and Yousef Hammad \at University of Illinois Urbana-Champaign, Urbana, Illinois, USA \\ \email{yhammad2@illinois.edu}
  \and Ahmed Samy \at Queen's University, Kingston, Ontario, Canada \\ \email{ahmed.samy@queensu.ca}
  \and Walid G. Aref~\orcidlink{0000-0001-8169-7775} \at Purdue University, West Lafayette, Indiana, USA \\ \email{aref@purdue.edu}
}

\date{Preprint (Under Review)}

\maketitle

\begin{abstract}
Large language models (LLMs) have transformed how humans interact with information, enabling natural-language interfaces across various domains. They are increasingly deployed for geospatial applications including map-based question answering, location-based services, and natural-language interfaces to spatial databases. But are LLMs truly \emph{location-aware}, able to reliably interpret and precisely answer location queries involving real-world coordinates, points of interest (POIs), and geospatial distances? To answer this concretely, we present {\bf LimoBench}, a street-level benchmark of 82 natural-language location queries over seven canonical spatial operations: geocoding, reverse geocoding, routing, directionality, range search, $k$-nearest-neighbor search, and spatial join. The queries are grounded in real map data of West Lafayette, Indiana, and are graded by location and semantic complexity. They are designed to probe LLMs' street-level spatial awareness, from simple lookups to multi-step spatial reasoning. We also present \textbf{Limo}, an approach based on the Model Context Protocol (MCP) that enhances the location awareness of LLMs. Limo grounds LLMs in PostGIS-backed spatial data through a compact set of tools. We find that ungrounded models mostly achieve single-digit accuracy (3.1\% for GPT-6 Luna) across different spatial operations. In contrast, the same models achieve near-perfect accuracy (99.3\% for GPT-6 Luna) when given access to Limo's spatial tools through MCP. Current LLMs thus have limited and unreliable street-level spatial knowledge, while grounding provides accurate answers at modest overhead (+0.2\,s latency, 7.7$\times$ tokens, and 2.2$\times$ cost for GPT-6 Luna). We release the benchmark, Limo toolset, and evaluation code to encourage the community to evaluate their own location-aware LLM approaches.
\end{abstract}

\begin{figure}[!t]
\centering
\includegraphics[width=0.99\linewidth]{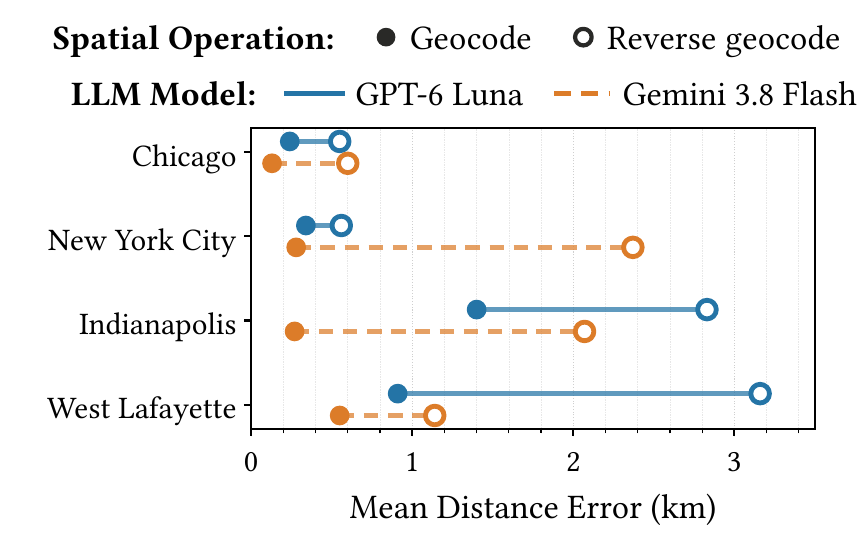}
\caption{Mean distance error of two LLMs on geocoding and reverse geocoding in four cities. Each point averages ten queries from a single run.}
\label{fig:motivation-study}
\end{figure}

\section{Introduction}
\label{intro}
Large language models (LLMs) provide effective interfaces to structured and unstructured data~\cite{li2024llmdatamgmt,gao2024texttosql,li2024dawnnl2sql}. Location-aware applications extend this interface to questions such as ``Which pharmacy is nearest to this address?'', ``How far is the trip by bicycle?'', and ``Which schools are within one kilometer of each neighborhood?'' Answering these questions requires both language understanding and precise operations over geographic data.

Street-level spatial queries expose a basic limitation of parametric knowledge. A model may associate a landmark with the correct city yet fail to place it within tens of meters, recover the nearby points of interest (POIs), or compute a route over the road network. Such errors are consequential because plausible coordinates and place names can be difficult for a user to distinguish from correct ones. Existing work likewise reports uneven geographic recall and spatial hallucination in LLMs~\cite{bhandari2023geospatially,wang2025mitigatinggeospatialknowledgehallucination,xie2025geospatial}. However, spatial databases and routing engines already implement these operations precisely. The open question is how well LLMs perform without them, and which tool abstractions allow models to use them reliably.

We start with {\bf geocoding} and {\bf reverse geocoding} because they are primitive but essential operations. Geocoding converts the place named by a user into a spatial reference, while reverse geocoding converts a coordinate back into a human-readable place. {\bf Routing}, {\bf directionality}, {\bf proximity search}, and {\bf spatial joins} commonly require one of these resolution steps before spatial computation can begin. Figure~\ref{fig:motivation-study} reports a preliminary ten-query probe per operation and model in four cities using GPT-6 Luna and Gemini 3.8 Flash, with low reasoning effort and temperature 0 for reproducibility. Even these primitive operations produce errors of hundreds of meters to several kilometers. Performance also varies by region, with West Lafayette and Indianapolis generally producing more errors than Chicago and New York City. This pattern is consistent with geographic data-availability bias in model pretraining, where less represented places are learned less precisely, and motivates explicit spatial grounding.

We address both of these questions. We introduce {\bf LimoBench}, a benchmark that evaluates street-level spatial awareness through seven operations drawn from spatial data management. Each operation contains ten location queries constructed from a common OpenStreetMap (OSM)~\cite{openstreetmap} snapshot. Queries are graded on two axes: location complexity, the number of spatial operations a query requires, and semantic complexity, the number of phrases that require interpretation beyond lexical matching against map data. The 70 location queries require one to five spatial operations and no semantic interpretation, and the 12 semantic queries add phrases that require it. The workload varies locations across the study region, geometry types, distance scales, travel modes, requested cardinalities, and join semantics. This controlled design separates coarse geographic familiarity from the ability to retrieve and compute exact spatial relationships.

Then, we introduce {\bf Limo}, a spatial Model Context Protocol (MCP) server that connects an LLM to a set of composable spatial tools. Place resolution tools identify places, geometry construction tools derive new spatial objects, routing and direction tools relate pairs of places, and search and join tools retrieve and relate sets of places. Tools exchange compact entity identifiers instead of repeating large geometries. Each tool performs one operation, reports validation failures explicitly, and accepts native OSM tag filters. Thus, the model selects and composes operations while established spatial systems execute them.

Our experiments compare ungrounded LLMs, model and inference ablations, Limo-grounded LLMs, and two competing spatial MCP servers. Every reported configuration contains three complete runs for each query set. With GPT-6 Luna, ungrounded accuracy is 3.1\%, and additional reasoning effort or sampling temperature does not materially resolve this limitation. Limo raises it to 99.3\% mean accuracy. Across a broader cross-provider model set, Limo attains 92.2--100.0\%, which indicates that the remaining variation primarily reflects tool-use ability. Competing MCP servers perform well on geocoding and reverse geocoding but degrade on routing, set retrieval, and joins, most sharply when the model must emulate a missing tool.

The contributions of this paper are as follows:
\begin{itemize}[leftmargin=*]
\item \textbf{LimoBench}, a street-level benchmark of 82 queries over seven canonical spatial operations, which measures the spatial awareness of LLMs at the granularity of individual streets, addresses, and POIs. Queries are graded by location complexity and semantic complexity, with 70 location and 12 semantic queries.
\item \textbf{Limo}, a portable MCP toolset that grounds LLMs in OSM data through 12 primitive and composable tools using PostGIS.
\item \textbf{An empirical study} of LLM-only and tool-grounded spatial querying across models, inference settings, and MCP servers, showing that grounding, not additional reasoning effort or sampling temperature, closes the accuracy gap.
\end{itemize}

We envision LimoBench as a living benchmark for location-aware LLM systems. Limo is a first step in this direction, and we invite the community to evaluate their LLM-based approaches on the same queries and ground truth.

The rest of the paper is organized as follows: Section~\ref{sec:background} introduces the necessary background. Section~\ref{sec:benchmark} presents LimoBench, and Section~\ref{sec:limo} describes the Limo spatial toolset. Section~\ref{sec:evaluation} presents the experimental evaluation. Section~\ref{sec:discussion} discusses the findings and limitations. Section~\ref{sec:related-work} reviews related work, and Section~\ref{sec:conclusion} concludes the paper.

\begin{table*}[!t]
\caption{Example LimoBench queries. Gray spans mark location components (L), underlined spans mark semantic components (S), and parentheses give query counts.}
\label{tab:benchmark-queries}
\centering
\begin{tabular}{
    l           
    >{\itshape\raggedright\arraybackslash}p{9cm} 
    l           
    p{2.3cm}    
}
\toprule
\textbf{Operation}
    & \textnormal{\textbf{Example}}
    & \textbf{Complexity}
    & \textbf{Answer} \\
\midrule
\multicolumn{4}{l}{\textbf{Location queries (70)}} \\
\midrule
Geocode
    & I want to find \hl{Universal Fitness, West Lafayette, IN} on a map. What coordinates should I use?
    & L1 & Coordinate \\
Reverse geocode
    & Someone dropped a pin at \hl{latitude 40.423297 and longitude -86.901085}. What's there?
    & L1 & Place or address \\
Directionality
    & From \hl{Taste of India}, \hl{what direction} would I face to look at \hl{Purdue Technology and Aerospace} in West Lafayette, IN?
    & L3 & Compass label \\
Route
    & \hl{By bike, how far} is it from \hl{Wabash Valley Education Center} to \hl{Unitarian Universalist Church of Tippecanoe County} in West Lafayette, IN?
    & L3 & Distance \\
Range
    & Take \hl{Cumberland West} and \hl{Covenant Presbyterian Church} as \hl{opposite corners of a rectangle} in West Lafayette, IN. What pharmacies lie inside it?
    & L3 & Place set \\
KNN
    & What are the \hl{5 nearest} supermarkets to \hl{Montessori School of Greater Lafayette} in West Lafayette, IN?
    & L2 & Ranked place set \\
Spatial join
    & Find every pair of \hl{a car repair shop} and \hl{a pub} \hl{in West Lafayette, IN} that are \hl{at most 200 meters apart}.
    & L4 & Place pairs \\
\midrule
\multicolumn{4}{l}{\textbf{Location + semantic queries (12)}} \\
\midrule
Geocode
    & Where do the \locsem{Boilermakers play their home football games}? I'd like coordinates I can drop a pin on.
    & L1S1 & Coordinate \\
Reverse geocode
    & My friend's phone only shows \locsem{``86.91365, 40.43011''} and she's somewhere in \loc{West Lafayette}. Which building is she at?
    & L1S1 & Place or address \\
Directionality
    & Standing at \locsem{B-Dubs}, in \loc{which compass direction} is \locsem{Purdue's basketball arena}?
    & L3S2 & Compass label \\
Route
    & \loc{By bike, how far} is it from \locsem{the barbershop next to Fresh Thyme} to \locsem{the hotel inside PMU}?
    & L5S3 & Distance \\
Range
    & Is there anywhere to \sem{catch a movie} \locsem{within a short bike ride of} \locsem{the vet school at Purdue}?
    & L2S3 & Place set \\
KNN
    & Can you find me \loc{a couple of} \sem{places to get a haircut} \loc{closest to} \locsem{the Co-Rec}?
    & L2S2 & Ranked place set \\
\bottomrule
\end{tabular}
\end{table*}

\section{Background}
\label{sec:background}

\subsection{Large Language Models}
LLMs are pretrained generative models that learn broad linguistic regularities and encode substantial world knowledge implicitly in their parameters (often described as parametric memory). Because this knowledge is not stored as explicit, queryable facts, purely parametric generation may produce hallucinations when the necessary knowledge is absent or weak~\cite{lewis2021retrievalaugmentedgenerationknowledgeintensivenlp}.

For geospatial tasks, parametric knowledge is especially brittle. Many real-world locations and place-specific facts are long-tail (rarely discussed in web text relative to major cities and landmarks), and long-tail coverage gaps correlate with poorer factual recall~\cite{kandpal2023largelanguagemodelsstruggle}. Empirically, geospatial hallucination benchmarks find that even advanced LLMs show low accuracy on structured geographic knowledge and that hallucinations are more pronounced in less represented regions, indicating uneven and unreliable geographic coverage from pretraining alone~\cite{wang2025mitigatinggeospatialknowledgehallucination}. Moreover, many location-aware queries require exact spatial computation (e.g., containment, distance ordering, shortest paths) with deterministic correctness guarantees, capabilities that do not naturally arise from next-token prediction and are better handled by specialized spatial systems.

These limitations motivate \emph{spatial grounding}. We use \emph{spatial grounding} to describe the process of anchoring model responses in external geographic data or spatial computation rather than relying solely on parametric knowledge. Under this paradigm, the LLM is responsible for interpreting user intent and composing actions, while authoritative results are produced by external systems such as geocoders, spatial databases, and routing engines. Prior work on tool-augmented language models has shown that delegating operations, such as arithmetic or factual retrieval, to external tools can substantially improve reliability, and data-management research has similarly argued for hybrid systems that combine LLM-based interpretation with database execution for queries that exceed the closed-world assumptions of stored tables~\cite{zhao2024hybridqueryingrelationaldatabases}. In our setting, spatial grounding is realized through MCP, which enables an LLM application to interact with external tools and data sources. This allows the system to support location-aware semantic querying while grounding spatial results in reliable computation over geographic data.

\subsection{Spatial Databases and Geographic Information Systems}
In this work, we use a PostgreSQL database with the PostGIS extension\footnote{https://postgis.net/} to execute spatial queries and to provide deterministic ground truth for evaluation. The benchmark follows standard spatial database conventions from the Open Geospatial Consortium~\cite{OGC_SFA_05_126} for representing geographic objects and evaluating spatial relationships, which provides a consistent basis for comparison.

Spatial databases and geographic information systems (GIS) represent geographic entities as geometries with associated attributes. They support core operations such as spatial predicates, range search, nearest-neighbor search, geocoding, reverse geocoding, and routing~\cite{goldberg2007,eldawy2017bigspatial,bakshi2004geocode,papadias1994qualitative,sommer2014shortest}. These operations correspond closely to many location-aware natural-language queries.

LLMs do not natively perform these operations with the precision, determinism, or data freshness required for reliable geographic reasoning~\cite{wang2025mitigatinggeospatialknowledgehallucination,xie2025geospatial}. For this reason, location-aware querying systems must ground LLM outputs in external spatial databases and services, which serve as the source of truth for spatial computation~\cite{lu2024gearaugmentinglanguagemodels,krechetova2025geobenchxbenchmarkingllmsagent}.

\subsection{Model Context Protocol}
MCP is an open protocol for connecting LLM applications to external data sources and tools through a common interface~\cite{anthropic_mcp_intro,mcp_docs_intro}. It reduces the need for a separate adapter between each model host and each external system. MCP follows a client-server architecture. An MCP server wraps some external resource or functionality, such as a database, filesystem, web API, or computation engine, and exposes it through the protocol. An MCP client, which runs inside an LLM-powered application, connects to one or more such servers and makes their capabilities available to the model~\cite{mcp_spec_schema}.

The protocol standardizes initialization, capability discovery, and message exchange. It uses JSON-RPC with transports including standard input/output (stdio) and HTTP-based communication~\cite{mcp_spec_transports}.

MCP supports runtime tool discovery: the client obtains tool names, descriptions, and input schemas from the server and passes them to the model. A tool-capable model can therefore use a new server without server-specific bindings or task-specific parameter updates, provided that the server's interface is clear.

In this work, we use MCP to expose spatial operations over a PostGIS spatial database. The model interprets the request and composes tool calls, while the spatial backend executes data access and computation.

\section{Benchmark Design}
\label{sec:benchmark}
LimoBench measures whether an LLM can answer precise street-level questions, not whether it can identify a city or region. It grades queries along two axes, location complexity and semantic complexity. It contains 82 queries: 70 location queries (ten for each of seven operations) and 12 semantic queries. We use West Lafayette, Indiana, as a city-scale test region and derive all locations, addresses, POIs, and road networks from a fixed OSM snapshot. The construction procedure is region-independent and can be applied to another OSM PBF extract.

\subsection{Spatial Operations}
In this benchmark, we consider seven spatial operations, defined below with example queries in Table~\ref{tab:benchmark-queries}.
\begin{itemize}[leftmargin=*]
    \item \emph{Geocoding} maps a place or an address to its coordinates.
    \item \emph{Reverse geocoding} retrieves the place or address associated with given coordinates.
    \item \emph{Directionality} assigns a compass direction from one location to another.
    \item \emph{Routing} measures travel distance between two locations along the road network.
    \item \emph{Range} queries retrieve the places within a reference geometry.
    \item \emph{$k$-nearest-neighbor (KNN)} queries retrieve the $k$ places closest to a reference geometry.
    \item \emph{Spatial joins} relate two sets of locations by distance or nearest-neighbor criteria.
\end{itemize}

\subsection{Query Complexity}
\label{sec:complexity}
A query becomes harder when it requires more spatial operations or when more of its wording must be interpreted before any operation can run. LimoBench grades these two sources separately.

\noindent\textbf{Location complexity (L).} L counts the spatial operations in the canonical execution plan of a query, which follows its ground-truth specification. Resolving each referenced place or region by geocoding or reverse geocoding is one operation. Each search, measurement, or join over resolved references is one more. Non-spatial post-processing, such as counting or sorting, is not counted. For example, a radius search around a named place is L2 (i.e., 1L for geocoding the named place + 1L for searching within the radius), and a route between two named places is L3 (i.e., 2L for geocoding the two named places + 1L for routing).

\noindent\textbf{Semantic complexity (S).} S counts the phrases that cannot be resolved by direct lexical lookup in a spatial database because the phrases are not part of the database's vocabulary and their meaning must be inferred. Conventional category nouns, such as \emph{pharmacy} or \emph{gas station}, map directly to OSM tags and are not counted. We distinguish three classes of semantic phrases:
\begin{itemize}[leftmargin=*]
    \item \emph{Anchor references} denote a place by a colloquial name, an abbreviation, or a description. \emph{PMU} denotes the Purdue Memorial Union, whose OSM name is \emph{Purdue Student Union}.
    \item \emph{Category references} denote a class of places by its function or attributes rather than by its OSM tag. \emph{Withdraw cash} resolves to \texttt{amenity=atm} or \texttt{amenity=bank}, and \emph{pick up medicine} to \texttt{amenity=pharmacy}.
    \item \emph{Proximity parameters} set $k$ or a distance threshold through vague quantities, such as \emph{a few} or \emph{walking distance}. Their resolution depends on context, including travel mode and distance scale.
\end{itemize}
A phrase counts once in S even when it also contributes a location component. Surface-level linguistic variation, such as typos, abbreviated street types, and spelled-out numbers, requires normalization but no world knowledge. We treat it as a lighter perturbation that can be applied at any level and do not count it in S.

\noindent\textbf{Complexity grid.} We write L$n$S$m$ for a query with $n$ location components and $m$ semantic components, and omit S when $m=0$, so L$n$ denotes a pure location query. The axes are independent because semantic phrases change how a query expresses its arguments (spatial predicates), not its execution plan. Every query has at least one location component, so S refines L and never replaces it. \emph{How far is the walk from PMU to the Co-Rec?} has the plan and answer of the L3 query that names the Purdue Student Union and France A. C\'ordova Recreational Sports Center, but is itself L3S2.

Table~\ref{tab:benchmark-queries} marks each component in its example queries. The 70 pure location queries have no semantic component and span L1 to L5. Geocoding and reverse geocoding are L1 (20 queries). KNN and radius range queries are L2 (15). Operations including directionality, routing, rectangle range queries, and one self-join are L3 (26). The remaining spatial joins are L4 (7) or L5 (2), depending on how many references and input sets they resolve. The 12 semantic queries span L1 to L5 and S1 to S3.

Semantic queries retain the requirement of one defensible answer. A semantic phrase qualifies when its plausible readings agree on the ground truth. For example, in the range query of Table~\ref{tab:benchmark-queries}, the nearest cinema to the vet school is far closer than the next one, so any reasonable reading of \emph{a short bike ride} returns the same place. Subjective qualities, such as \emph{family-friendly}, and vernacular regions, such as \emph{downtown}, have no such stable resolution and are excluded.

\subsection{Workload Construction}
The workload applies four design principles. First, it covers the main spatial objects and operators used in location services. References include points, building polygons, and bounding regions. Locations are distributed across the study area to avoid concentrating every query around the university campus. Second, it controls difficulty within each operation by varying travel profiles, route length, $k$, distance type, reference geometry, set size, and join type (Table~\ref{tab:benchmark-stats}). Third, paired operations remain internally consistent. Each geocoding query has a corresponding reverse-geocoding instance, and directionality queries cover the eight compass sectors at different distances. Fourth, multiple natural-language templates vary surface form while preserving executable semantics.

These principles increase the diversity of benchmark queries and reduce bias toward one geometry, distance scale, or answer size. They also parameterize benchmark generation by region and requested workload size. We retain ten instances per operation to control API cost because the workload is repeated across runs, models, inference settings, and MCP servers. The benchmark ships natural-language questions, structured metadata, and database-derived ground truth. Ground truth is never included in the model context.

\subsection{Ground Truth}
We import the OSM extract of West Lafayette (35.7~km$^2$ and 494 POIs in 116 categories) with \texttt{osm2pgsql}\footnote{https://osm2pgsql.org/} into PostgreSQL with PostGIS. Photon\footnote{https://github.com/komoot/photon}, an open-source geocoding tool, supplies consistent forward and reverse geocoding over the same extract, and GraphHopper\footnote{https://www.graphhopper.com/open-source/}, an open-source routing engine, supplies routes for the three travel profiles. Generation retains only instances with valid inputs and executable answers, and each query has a single database-derived answer. Ground-truth records preserve OSM entity identifiers where available, which avoids ambiguous name-only matching. Spatial joins additionally record whether their pairs are unordered, as in self-joins, or directed, as in nearest-per-left joins. Table~\ref{tab:benchmark-stats} summarizes the resulting queries.

\begin{table}[t]
  \centering
  \small
  \caption{LimoBench statistics over all 82 queries.}
  \label{tab:benchmark-stats}
  \begin{tabular}{@{}ll@{}}
    \toprule
    \textbf{Statistic}           & \textbf{Value} \\
    \midrule
    Queries                      & 82 (70 location, 12 semantic) \\
    Distinct ground-truth places & 162 (63 points, 99 polygons) \\
    Tag filters                  & 24 (amenity 16, shop 6, etc.) \\
    Area covered                 & 5.8 $\times$ 4.9 km \\
    Complexity levels            & L1 to L5, S0 to S3 \\
    Route length                 & 0.8 to 6.4 km \\
    Range radius                 & 0.4 to 3 km \\
    Join distance                & 100 to 300 m \\
    KNN $k$                      & 1 to 8 \\
    Answer set size              & 1 to 10 \\
    \bottomrule
  \end{tabular}
\end{table}

\section{Limo Spatial Toolset}
\label{sec:limo}
\begin{figure*}[!t]
\centering
\includegraphics[width=0.99\linewidth]{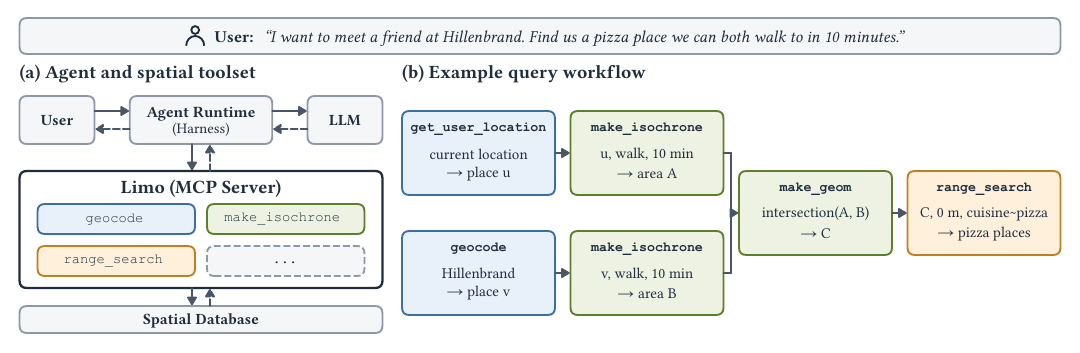}
\caption{Limo overview. (a) The agent runtime calls Limo's tools through MCP, and the tools query a spatial database. (b) Tool workflow for the example query: two isochrones are intersected and searched for pizza places.}
\label{fig:limo-overview}
\end{figure*}

Limo exposes spatial data and computation to an LLM through MCP. Given an OSM PBF extract, its build pipeline imports geometries and tags into PostGIS with \texttt{osm2pgsql} and constructs matching Photon and GraphHopper indexes. Photon supplies forward and reverse geocoding, while GraphHopper supplies car, bicycle, and pedestrian routing. Reusing these established services avoids reimplementing mature spatial functionality. MCP hides the database and services behind JSON tool schemas that any compatible model host can discover at inference time. Tool outputs also retain geometries and routes for visualization by a compatible client. Figure~\ref{fig:limo-overview} shows how the agent runtime connects the LLM to Limo and how the tools compose for an example query.

\subsection{Design Principles}

\noindent\textbf{Single responsibility.} Each tool performs one spatial operation. For example, \texttt{range\_search} accepts a resolved reference and does not silently geocode a name. A failed resolution is therefore visible to the model and can be corrected before a later operation returns a plausible result for the wrong place.

\noindent\textbf{Composition.} Tools form four groups (Table~\ref{tab:limo-tools}). Place resolution tools look up and geocode places. Geometry construction tools build derived geometries and isochrones. Routing and direction tools compute routes and bearings between places. Search and join tools retrieve places by range or nearest neighbor and relate two retrieved sets. This design supports workflows not encoded as monolithic tools. For example, the model can geocode an address, construct a walking isochrone, search for pharmacies inside it, and route to the selected result.

\noindent\textbf{Compact state.} Every returned place or derived geometry has an \texttt{entity\_id}. Later calls pass these identifiers instead of serializing geometries through the model context. This reduces token use, preserves the full geometry, and allows points, lines, polygons, routes, and isochrones to share the same reference interface.

\noindent\textbf{Uniform interfaces.} Tools use common place references, units, filters, limits, and location objects. The same search interface accepts a coordinate, a building polygon, a street line, or a derived geometry. Concise and consistent schemas reduce interface-specific decisions left to the model.

\noindent\textbf{OSM-native filtering.} Search tools filter native OSM tags with a compact Boolean grammar. They support equality, alternatives, tag presence, substring matching, negation, conjunction, and disjunction. Retaining native tags avoids a lossy application-specific POI taxonomy and makes the server portable across OSM extracts.

\noindent\textbf{Explicit guidance.} Inputs are validated for coordinate ranges, identifier existence, set size, and incompatible options. Results echo interpreted filters and include concise guidance for likely next steps. Errors state how to repair a call, which helps models recover without concealing failures.

\begin{table*}[!t]
\caption{Limo tools by group. \texttt{PlaceRef} is an entity identifier or a WGS84 coordinate.}
\label{tab:limo-tools}
\centering
\small
\begin{tabular}{ll p{4.3cm} p{5.5cm}}
\toprule
\textbf{Tool Group} & \textbf{Tool} & \textbf{Input} & \textbf{Output} \\
\midrule
\multirow{4}{*}{\tg{Place Resolution}{blue}} & \texttt{lookup} & Identifier & One place \\
 & \texttt{geocode} & Name or address & Ranked places with entity identifiers \\
 & \texttt{reverse\_geocode} & Coordinate & Nearest places \\
 & \texttt{get\_user\_location} & None & Device location and accuracy \\
\midrule
\multirow{2}{*}{\tg{Geometry Construction}{green}} & \texttt{make\_geom} & \texttt{PlaceRef}s, geometry operator & Derived geometry entity \\
 & \texttt{make\_isochrone} & \texttt{PlaceRef}, travel mode, time or distance budget & Reachable-area entity \\
\midrule
\multirow{2}{*}{\tg{Routing \& Directions}{teal}} & \texttt{route} & Two \texttt{PlaceRef}s, travel mode & Route, distance, duration, and summary \\
 & \texttt{bearing} & Two \texttt{PlaceRef}s & Initial bearing and compass direction \\
\midrule
\multirow{4}{*}{\tg{Search \& Join}{orange}} & \texttt{knn\_search} & \texttt{PlaceRef}, $k$, OSM filter & Nearest places by geodesic distance \\
 & \texttt{range\_search} & \texttt{PlaceRef}, radius, OSM filter & Places within the range \\
 & \texttt{knn\_search\_network} & \texttt{PlaceRef}, $k$, travel mode, filter & Nearest places by travel distance \\
 & \texttt{spatial\_join} & Two entity sets, join constraints & Place pairs or selected left places \\
\bottomrule
\end{tabular}
\end{table*}

\subsection{Tools}
Table~\ref{tab:limo-tools} lists the 12 tools used in the study. Geodesic searches compute distance between full geometries. Network searches and routes use GraphHopper travel profiles. \texttt{spatial\_join} supports inner, left, semi, and anti joins, an optional distance threshold, and an optional nearest-match limit per left input. A dedicated join tool is not strictly required for expressiveness, but it avoids repeated model-controlled searches and distance computations.

\subsection{Why Specialized Spatial Tools Matter}
General database access can make a server superficially expressive, but it transfers schema discovery, SQL generation, spatial semantics, and error recovery to the model. The a-tsitanov server~\cite{tsitanov_osm_mcp}, for example, exposes SQL execution that can fill some coverage gaps but requires the model to construct the operation correctly. Limo instead fixes the invariant parts of spatial execution in tested tools while leaving intent resolution and plan composition to the LLM. The MCP server comparison in Section~\ref{sec:results} evaluates this distinction. Competing servers perform well on directly supported primitive operations, yet lose accuracy and efficiency when the model must emulate routing, KNN, range, or join behavior through lower-level calls.

\section{Experimental Evaluation}
\label{sec:evaluation}

\subsection{Setup}
We evaluate two conditions. The \emph{LLM-only} condition gives the question directly to the model without tools. The \emph{grounded} condition gives the same question to an MCP agent.

The experiments are run on a single Apple M1 machine (arm64, 8 cores, 8~GB of memory) with macOS 15.6. The PostGIS, Photon, and GraphHopper services run locally on this machine, while LLM inference runs remotely through the provider APIs. The reference configuration uses GPT-6 Luna, low reasoning effort, temperature 0, and seed 42. Low effort limits cost while retaining tool-use ability. We study reasoning effort, temperature, models within the OpenAI family, models across providers, and three MCP servers. We use OpenRouter\footnote{https://openrouter.ai/} as a unified model gateway for all non-Claude models so that the experiments share one API interface. Claude models use the direct Anthropic API. We request low reasoning and temperature 0 unless stated otherwise. GPT-4o and Qwen3 Coder do not expose the reasoning setting used by the other models, so their runs omit it. Direct Anthropic requests omit temperature because the evaluated Claude endpoint rejects values below 1. In total, the cost of LLM API calls for running all the experiments is about \$90, of which the Claude models account for roughly two-thirds, mainly Opus and Sonnet. In contrast, the reference model GPT-6 Luna is lightweight, and all its runs together cost about \$1.50.

Each configuration contains three independent benchmark runs of every query set it covers. Unless specified otherwise, results use the 70 location queries. Section~\ref{sec:semantic-results} evaluates Limo on the 12 semantic queries. Warm-up queries are excluded. Failed and unanswered queries remain in all accuracy, latency, and cost denominators.

The MCP server comparison includes NERVsystems/\allowbreak osmmcp~\cite{nervsystems_osmmcp}, which calls Nominatim~\cite{nominatim}, Overpass~\cite{overpass_api}, and OSRM~\cite{luxen2011osrm}, and a-tsitanov/\allowbreak osm\_\allowbreak mcp~\cite{tsitanov_osm_mcp}, which queries a PostGIS database imported with \texttt{osm2pgsql}. Public Nominatim and Overpass endpoints are rate limited, so for a fair comparison we point NERVsystems/\allowbreak osmmcp at self-hosted instances and run all systems locally on the same OSM extract. The servers differ in tool coverage, output schemas, and backing services. We therefore interpret this comparison as an end-to-end evaluation of grounding designs. Table~\ref{tab:mcp-support-current} summarizes dedicated operation support in the MCP servers. A missing tool does not imply that an operation is impossible. It means that the model must approximate it by composing lower-level tools, issuing a general database query, or answering without grounded execution.

\begin{table}[!t]
\caption{Operation support across the evaluated MCP servers ($\checkmark$ = dedicated tool, $\times$ = no dedicated tool).}
\label{tab:mcp-support-current}
\centering
\begin{tabular}{lccc}
\toprule
\textbf{Operation} & \shortstack{\textbf{Limo}\\(ours)} & \shortstack{\textbf{NERV}\\\cite{nervsystems_osmmcp}} & \shortstack{\textbf{a-tsitanov}\\\cite{tsitanov_osm_mcp}} \\
\midrule
Geocode         & \checkmark    & \checkmark   & \checkmark \\
Reverse geocode & \checkmark    & \checkmark   & \checkmark \\
Route           & \checkmark    & \checkmark   & $\times$ \\
Directionality  & \checkmark    & $\times$     & $\times$ \\
Range           & \checkmark    & \checkmark   & \checkmark \\
KNN             & \checkmark    & $\times$     & \checkmark \\
Spatial join    & \checkmark    & $\times$     & $\times$ \\
\bottomrule
\end{tabular}
\end{table}

\begin{figure*}[!t]
\centering
\includegraphics[width=0.99\linewidth]{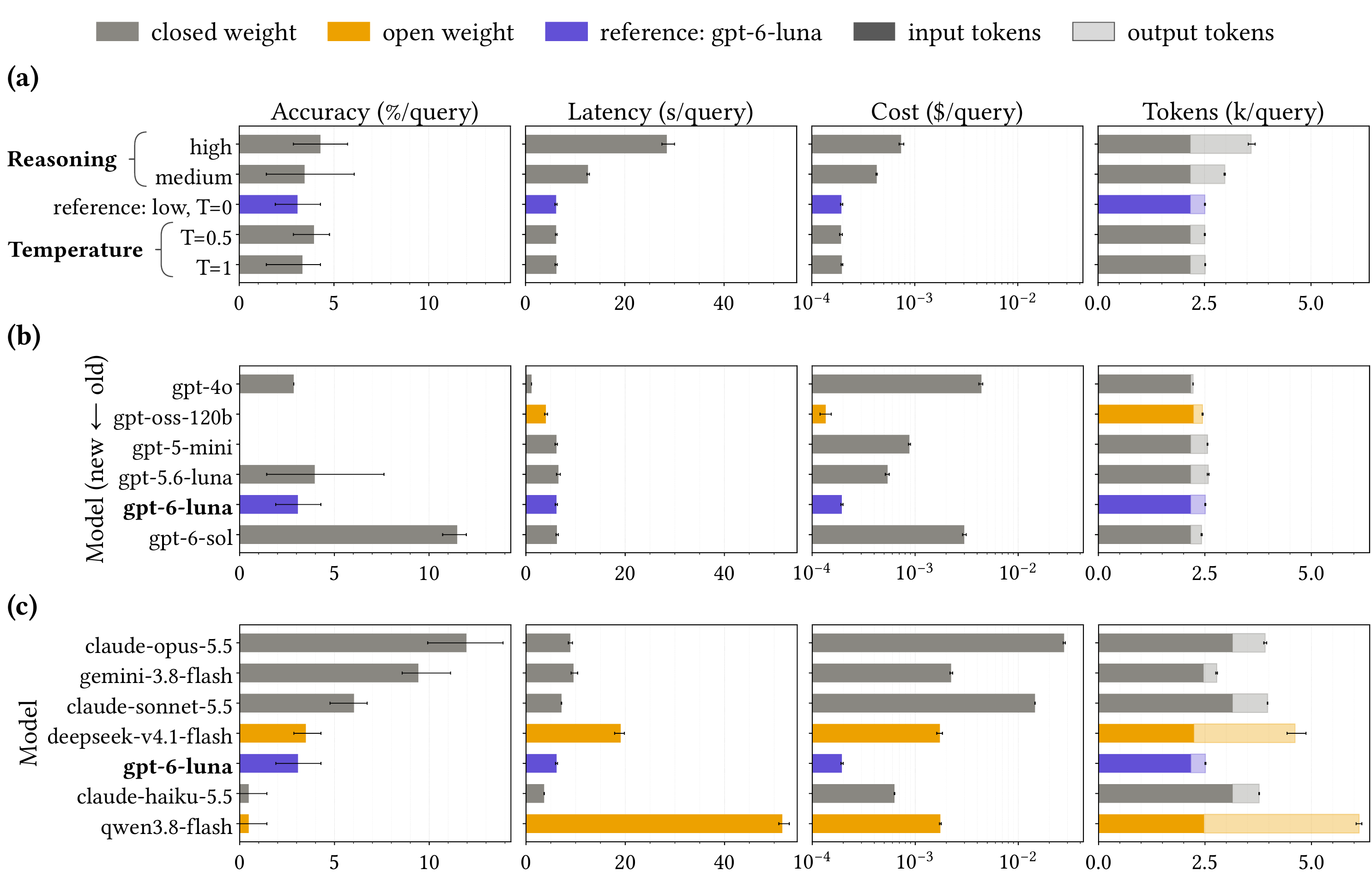}
\caption{LLM-only results for (a) reasoning effort and temperature with GPT-6 Luna, (b) OpenAI models, and (c) model providers.}
\label{fig:baseline-ablation}
\end{figure*}

The LLM-only and grounded system prompts share task semantics, formatting requirements, and one example for each operation. These examples standardize structured outputs for the evaluator. The grounded prompt additionally instructs the model to select tools, inspect their results, and reason carefully before answering. Ground truth remains outside both prompts.

\subsection{Metrics}
We report \emph{accuracy} as the mean per-query score. Geocoding is correct when its coordinate lies within 50 meters of ground truth. Reverse geocoding requires the correct place or address identity. Directionality requires the exact eight-point compass label. A route is correct when its distance is within 1\% of ground truth or differs by at most one meter. Range, KNN, and spatial join use membership F1 between the predicted and ground-truth sets. Join comparison respects whether pairs are directed or unordered. Semantic queries use the metric of their operation. Missing or invalid answers score zero. We further classify each answer as correct (full score), partially correct (F1 strictly between 0 and 1 for range, KNN, or spatial join), incorrect (valid but scoring zero), or invalid (no recoverable result).

We also report mean end-to-end latency per query, provider-reported cost, input and output tokens, and tool calls. Direct Anthropic costs are estimates and are not compared as provider-reported values. A failed tool call is an error or timeout. A repeated call is a successful call with the same tool and arguments as an earlier successful call in the same query. In bar figures, bars report the mean over the three runs and whiskers span their minimum and maximum.

\subsection{Results}
\label{sec:results}

\subsubsection{LLMs Without Spatial Tools}

\noindent\textbf{RQ1: Do reasoning effort or temperature improve street-level accuracy?} Figure~\ref{fig:baseline-ablation}(a) shows that increasing reasoning from low to high changes GPT-6 Luna accuracy from 3.1\% to 4.3\%, while latency rises from 6.18 to 28.47 seconds and cost rises from \$0.00019 to \$0.00074 per query. Temperature settings produce 3.1--3.9\% accuracy with nearly unchanged latency and cost. These small descriptive differences fall within an already unusable accuracy range. Neither setting supplies missing street-level facts or spatial computation.

\noindent\textbf{RQ2: How does spatial awareness vary across model generations?} Within the OpenAI family, newer models generally improve accuracy, but the trend is not monotonic. GPT-6 Sol is best at 11.5\%, while GPT-5 Mini and GPT-OSS 120B score 0\% in Figure~\ref{fig:baseline-ablation}(b). Newer models do not consistently reduce latency or cost because flagship and compact variants target different operating points.

\noindent\textbf{RQ3: How does performance vary across model families?} The cross-provider LLM-only configurations in Figure~\ref{fig:baseline-ablation}(c) span 0.5--12.0\% mean accuracy. Claude Opus 5.5 is highest at 12.0\%, but this remains insufficient for street-level applications. Open-weight Qwen and DeepSeek configurations use substantially more time or tokens than most closed-weight models without a corresponding accuracy advantage. GPT-6 Luna reaches 3.1\% and Gemini 3.8 Flash reaches 9.4\%. Directionality is the only operation with consistent nonzero exact accuracy, at 20.0\% and 40.0\% for these two models. Models sometimes recover coarse compass relationships but remain unreliable for precise retrieval and computation.

\begin{figure}[!t]
\centering
\includegraphics[width=0.99\linewidth]{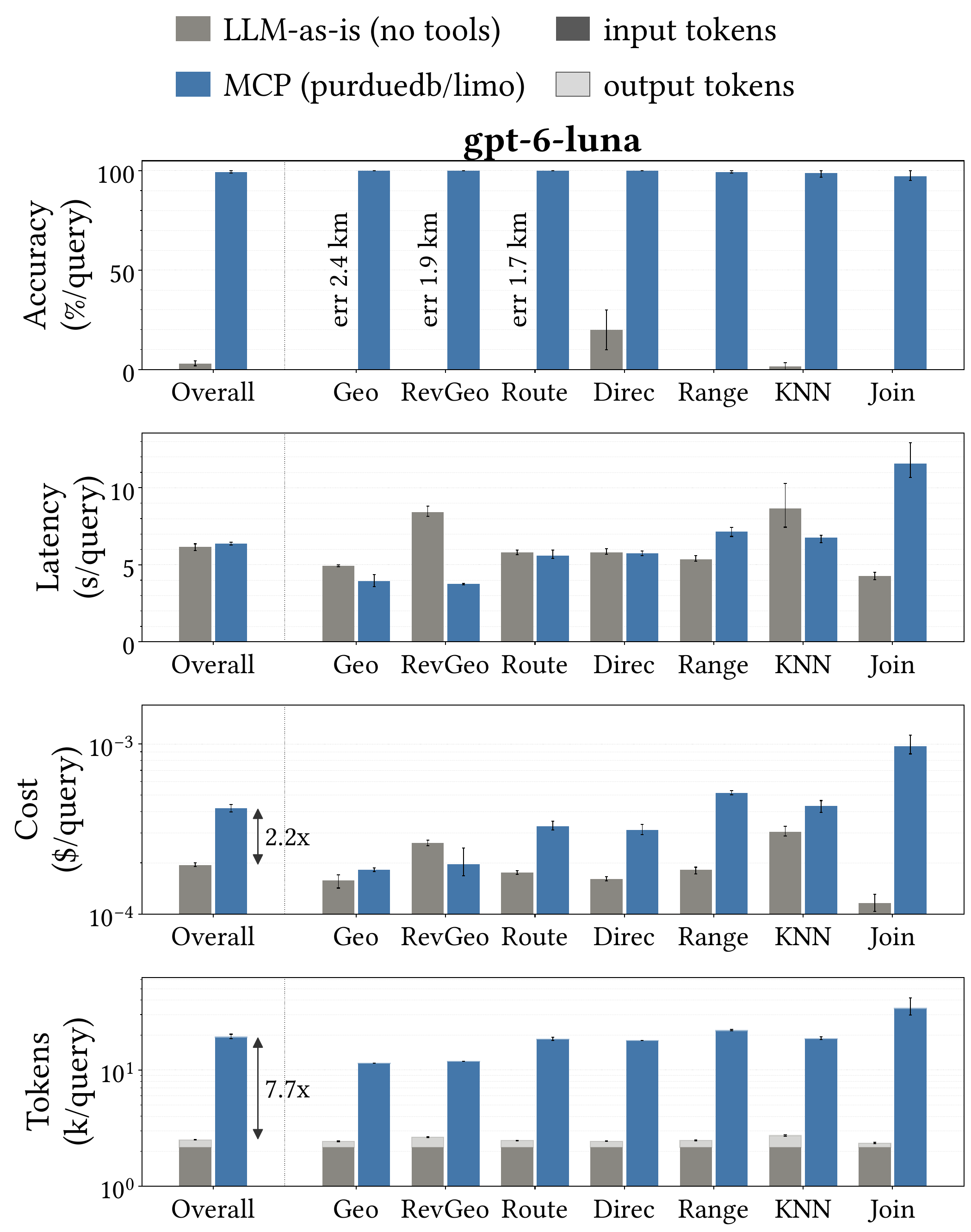}
\caption{GPT-6 Luna with and without Limo, by operation. The err labels give the LLM-only mean distance error over answered queries for geocoding, reverse geocoding, and routing.}
\label{fig:grounding}
\end{figure}

\subsubsection{Effect of Limo Grounding}

\noindent\textbf{RQ4: How much does spatial grounding improve performance?} Figure~\ref{fig:grounding} compares GPT-6 Luna with and without Limo. Its accuracy improves from 3.1\% to 99.3\%. Limo reaches 100\% on geocoding, reverse geocoding, routing, and directionality with GPT-6 Luna. Its remaining mean scores are 99.3\% for range, 98.9\% for KNN, and 97.2\% for spatial join. The lower join score reflects errors in the first step, collecting the input sets. In one query, two of three runs matched a category phrase by name instead of by tag, and thus propagated this error to the final result.

The mean latency changes from 6.18 to 6.36 seconds per query. Thus, the reference Limo runs add little end-to-end latency relative to generation. Cost rises by a factor of 2.2 and total tokens by a factor of 7.7 because every model turn receives tool schemas and tool results. Cost grows less than token count because much of the added context is input, while the LLM-only condition spends a larger share on output and reasoning tokens.

\begin{figure*}[!t]
\centering
\includegraphics[width=0.99\linewidth]{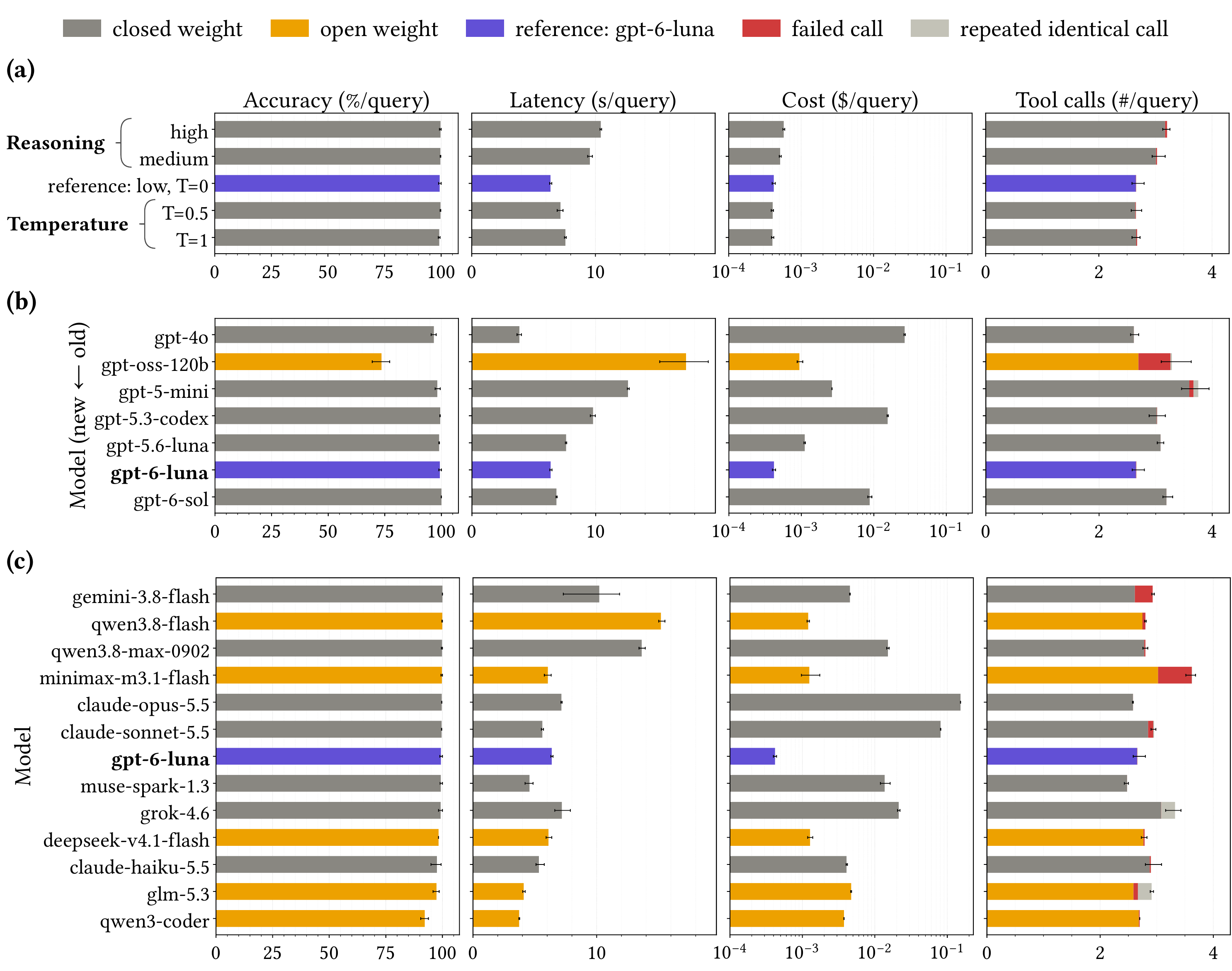}
\caption{Limo results for (a) reasoning effort and temperature with GPT-6 Luna, (b) OpenAI models, and (c) model providers.}
\label{fig:mcp-ablation}
\end{figure*}

\subsubsection{Inference and Model Ablations with Limo}

\noindent\textbf{RQ5: Do reasoning effort or temperature improve grounded performance?} Reasoning effort and temperature have limited effect once tools perform the spatial computation, as shown in Figure~\ref{fig:mcp-ablation}(a). Low, medium, and high effort yield 99.3\%, 99.7\%, and 99.7\% accuracy. High effort increases latency from 6.36 to 10.43 seconds and cost from \$0.00042 to \$0.00057 per query. Temperature variants remain between 99.1\% and 99.7\%. The reference tasks mainly require selecting a tool, supplying validated arguments, and formatting its result. Accuracy is already near its ceiling, so the added latency of higher effort is not justified for this workload.

\noindent\textbf{RQ6: How has grounded tool use changed across model generations?} The OpenAI family results in Figure~\ref{fig:mcp-ablation}(b) show that newer models generally invoke tools more reliably than older or smaller models. GPT-6 Sol reaches 100.0\%, while open-weight GPT-OSS 120B reaches 73.6\% with 0.57 failed calls per query. The trend is not monotonic because flagship, compact, and open-weight variants differ in intended operating point. For example, the older GPT-4o is the fastest family member at 3.83 seconds per query but also the most expensive. Failed and repeated calls expose cases where a model submits invalid arguments or retries an identical successful call.

\noindent\textbf{RQ7: How does grounded performance vary across model families?} The broader comparison in Figure~\ref{fig:mcp-ablation}(c) spans 92.2--100.0\% mean accuracy with Limo. Gemini 3.8 Flash reaches 100.0\%, but averages 10.20 seconds and \$0.00453 per query. MiniMax has the largest failed-call rate, 0.60 calls per query, yet recovers to 99.7\% accuracy. GPT-6 Luna reaches 99.3\% at 6.36 seconds and \$0.00042. It provides the best balance for the reference configuration, although an application's preferred model depends on its accuracy, latency, and cost constraints.

\begin{figure}[!t]
\centering
\includegraphics[width=0.99\linewidth]{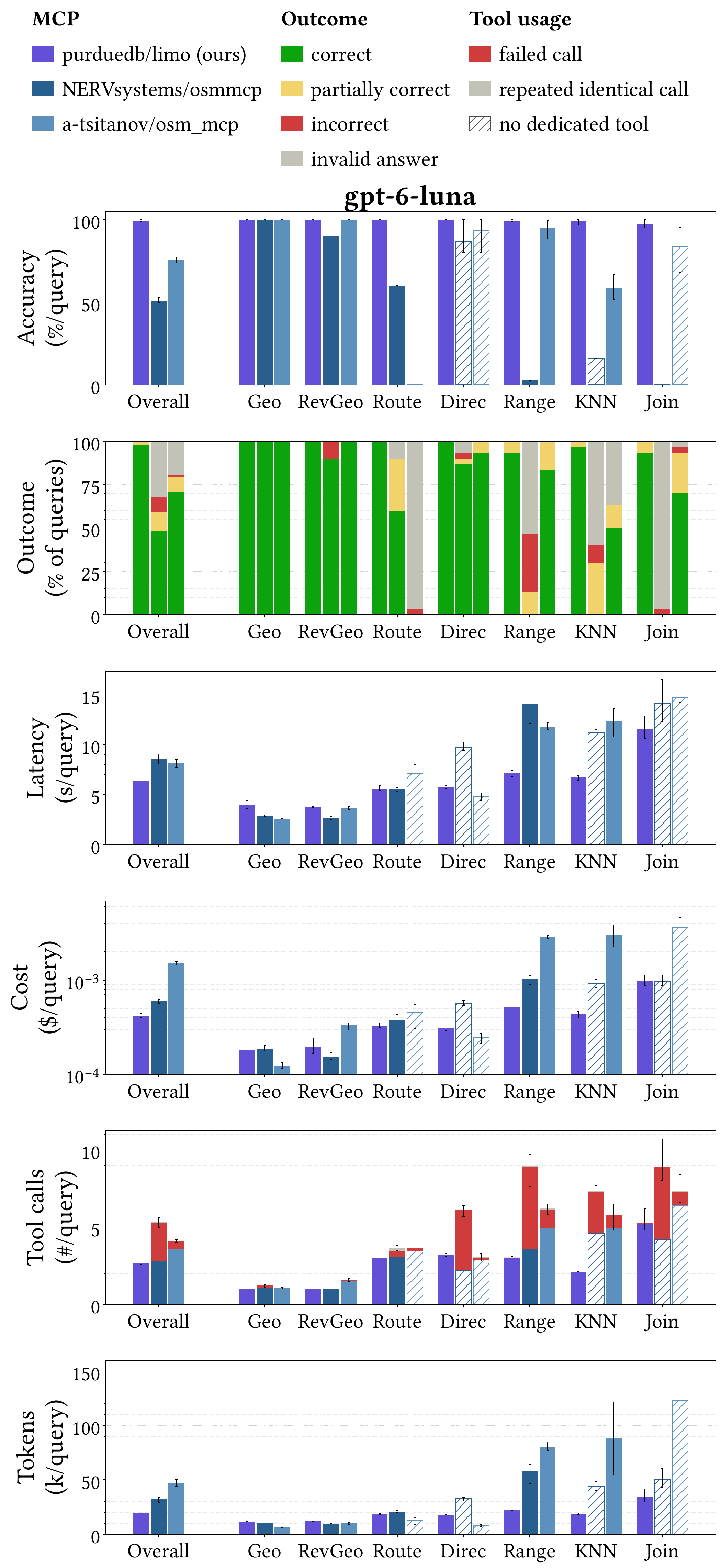}
\caption{Comparison of Limo and other spatial MCP servers with GPT-6 Luna, by operation. Hatched bars mark operations without a dedicated tool.}
\label{fig:mcp-methods}
\end{figure}

\subsubsection{Grounding Toolset Comparison}

\noindent\textbf{RQ8: How does performance vary across spatial grounding methods?} Figure~\ref{fig:mcp-methods} compares complete grounded systems. With GPT-6 Luna, Limo reaches 99.3\%, NERV reaches 50.8\%, and a-tsitanov reaches 75.8\%. The operation breakdown explains the gap. Missing dedicated tools cause sharp losses on structured operations. For example, a-tsitanov has no routing tool and scores 0.0\% with GPT-6 Luna. NERV has no dedicated spatial join tool and scores 0.0\% on that operation with GPT-6 Luna. The outcome breakdown in the same figure further distinguishes incomplete sets from total failures.

Lower-level access does not fully compensate for missing operations. Models attempt multi-call workarounds, which increase tokens, failed calls, and latency while leaving more decisions to generation. Limo averages 2.66 tool calls and 19.2 thousand tokens per query with GPT-6 Luna. NERV averages 5.32 calls and 32.2 thousand tokens, while a-tsitanov averages 4.10 calls and 46.9 thousand tokens. Specialized, composable tools therefore improve both correctness and execution efficiency.

\subsubsection{Semantic Queries}
\label{sec:semantic-results}

\noindent\textbf{RQ9: How well does Limo answer semantic queries?} The 12 semantic queries add anchor, category, and proximity phrases to the location operations (Table~\ref{tab:benchmark-queries}). The model must map each phrase to an OSM name, a tag predicate, or a parameter value before the corresponding spatial tool can run. Table~\ref{tab:semantic-results} reports three runs per effort. Semantic queries are more complex than location queries, so accuracy drops slightly to 95.8\% at low and 97.2\% at medium effort, while high effort reaches 100\%. The 3 misses among 108 task-runs involve no tool or API failures. Two directionality answers anchor on the wrong Starbucks next to a Chipotle, and one range answer adds a shipping place outside the accepted set. The model averages 5.03 to 7.89 tool calls per query, 89\% to 146\% more than on location queries, because it first resolves each described anchor or category to an OSM entity or tag. These extra calls raise latency by 16\% to 65\% and cost by 71\% to 119\%, so high effort removes the misses at about 2.2 times the location cost. With only 12 queries and 3 runs, accuracy differences between efforts amount to a few task-runs.

\begin{table}[!t]
\caption{Limo with GPT-6 Luna on the 12 semantic queries, by reasoning effort. Parentheses: change from the 70 location queries, with accuracy in percentage points (pp).}
\label{tab:semantic-results}
\centering
\footnotesize
\setlength{\tabcolsep}{3.5pt}
\begin{tabular}{
    l 
    r 
    r 
    r 
}
\toprule
\textbf{Metric}
    & \textbf{Low}
    & \textbf{Medium}
    & \textbf{High} \\
\midrule
Accuracy (\%)                 & 95.8 ($-$3.5\,pp) & 97.2 ($-$2.4\,pp) & 100 (+0.3\,pp) \\
Latency (s)                   & 8.65 (+36\%)    & 11.08 (+16\%)   & 17.26 (+65\%) \\
Cost ($10^{-3}$\,\$)          & 0.72 (+71\%)    & 0.91 (+76\%)    & 1.26 (+119\%) \\
Tool calls                    & 5.03 (+89\%)    & 5.89 (+94\%)    & 7.89 (+146\%) \\
\bottomrule
\end{tabular}
\end{table}

\begin{figure}[!t]
\centering
\includegraphics[width=0.99\linewidth]{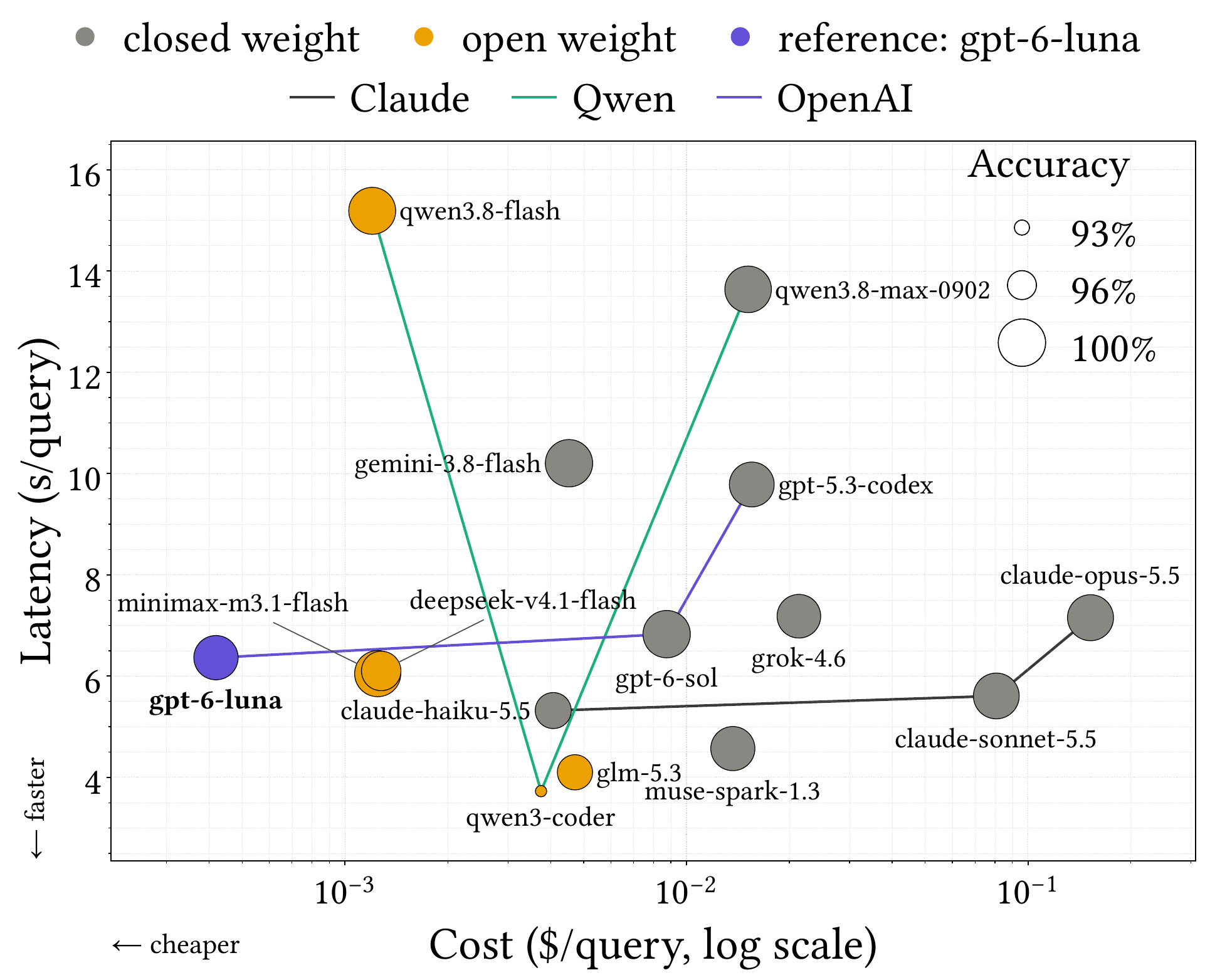}
\caption{Cost (log scale) versus latency for Limo model configurations. Marker size denotes accuracy, and lines connect models of one family.}
\label{fig:tradeoff}
\end{figure}

\section{Discussion}
\label{sec:discussion}
Figure~\ref{fig:tradeoff} summarizes the practical accuracy, latency, and cost trade-off across Limo model configurations. The lower-left region contains faster and cheaper models, while marker size indicates accuracy. GPT-6 Luna occupies a favorable position, but no model dominates every criterion. The preferred configuration therefore depends on the application's accuracy requirements and resource constraints.

\noindent\textbf{Parametric geographic knowledge is approximate.} Models often return plausible regional answers, but near-zero exact performance on reverse geocoding, range, KNN, and joins, and at most 13.3\% on geocoding and routing, shows that this knowledge does not support dependable street-level queries. More inference-time reasoning cannot recover facts and computations absent from the context.

\noindent\textbf{Tool design determines agent reliability.} MCP provides a common transport and discovery layer, not a guarantee of useful abstractions. Results improve when tools align with spatial operators, validate inputs, preserve geometry through identifiers, and compose without duplicating responsibility. A generic SQL tool increases theoretical coverage but also asks the model to perform database engineering within each answer.

\noindent\textbf{Grounding changes the model's role.} With Limo, the LLM interprets intent and constructs a short spatial plan. PostGIS, Photon, and GraphHopper supply data access and numerical correctness. The small benefit from higher reasoning effort follows from this division of labor. Queries with higher location or semantic complexity may require stronger planning.

\noindent\textbf{Limitations.} LimoBench contains ten location queries per operation and 12 semantic queries in one city. Three repetitions characterize observed variability but do not establish geographic generalization or statistical significance. OSM coverage and tags vary by region. The compared MCP servers use different services and data layouts, so the server comparison combines tool design with backend effects. Provider serving conditions, caches, and pricing also vary. Finally, the semantic queries exclude subjective concepts such as ``quiet'' and vernacular regions such as ``downtown'' because they lack defensible ground truth.

\section{Related Work}
\label{sec:related-work}
LLM geospatial research includes prompt-only evaluation, task-specific fine-tuning, spatial pretraining, modified objectives, and tool grounding~\cite{hussein2025spatialqueries}. Prompt-only studies measure geographic knowledge already encoded in model parameters~\cite{bhandari2023geospatially}. Spatially adapted models learn representations for entities, trajectories, and urban processes~\cite{li2022spabertpretrainedlanguagemodel,li2023geolmempoweringlanguagemodels,li2024urbangptspatiotemporallargelanguage}. Geo2Vec~\cite{chu2026geo2vec} learns unified embeddings of points, polylines, and polygons that encode shape, location, and topological and distance relations. These approaches complement tool grounding. They improve learned spatial structure but do not guarantee current database contents or exact execution.

Retrieval and tool augmentation instead delegate factual access or computation to external systems~\cite{lewis2021retrievalaugmentedgenerationknowledgeintensivenlp,zhao2024hybridqueryingrelationaldatabases}. MCP standardizes runtime discovery and invocation of such tools~\cite{anthropic_mcp_intro,mcp_docs_intro}. Recent geospatial agents likewise connect models to GIS functions or data sources~\cite{lu2024gearaugmentinglanguagemodels,krechetova2025geobenchxbenchmarkingllmsagent}.

Open-source spatial MCP servers differ in data source, operation coverage, output format, and tool granularity. PostGIS-backed servers query a local database. The general-purpose \mbox{receptopalak}/\allowbreak postgis-mcp~\cite{receptopalak_postgis_mcp} exposes tools that wrap geometry construction, measurement, overlay, and SQL queries. These primitives are flexible but not specific to OSM, so the model must discover the schema and assemble higher-level workflows such as geocoding itself. The OSM-specific a-tsitanov/\allowbreak osm\_\allowbreak mcp~\cite{tsitanov_osm_mcp} targets the \texttt{osm2pgsql} schema with feature search, proximity, area, and SQL tools, whereas \mbox{wiseman}/\allowbreak osm-mcp~\cite{wiseman_osm_mcp} pairs one PostGIS query tool with Leaflet map controls and focuses on visualization. Service-backed servers call OSM web services. \mbox{NERVsystems}/\allowbreak osmmcp~\cite{nervsystems_osmmcp} combines Nominatim, Overpass, and OSRM with composable primitives and composite tools for neighborhood and commute analysis. Other servers offer geocoding, nearby or bounding-box search, and raw Overpass queries~\cite{cyanheads_openstreetmap_mcp,neco001_openstreetmap_mcp}, or only geocoding and distance~\cite{webcoderz_mcp_geo}. General Postgres servers expose schema inspection and SQL without spatial abstractions~\cite{anthropic_postgres_mcp,crystaldba_postgres_mcp,deltorro2_geo_post_mcp}. Proprietary map tools provide geocoding, place search, and routing over closed data~\cite{anthropic_google_maps_mcp,cablate_mcp_google_map,google_maps_grounding,mapbox_mcp_server}, so their answers cannot be fairly compared against an open ground-truth database.

These projects are under active development and make spatial data readily accessible to LLM agents. To our knowledge, none is accompanied by a peer-reviewed evaluation, and no prior work benchmarks LLMs with a spatial grounding toolset at street-level granularity. Our work complements these efforts with a benchmark that measures how well such servers support canonical spatial operations and how tool granularity affects accuracy and efficiency.

\section{Conclusion}
\label{sec:conclusion}
LimoBench tests street-level spatial awareness through 82 queries, 70 location and 12 semantic, over seven spatial operations. Across models and inference settings, LLM-only execution remains unreliable, with single-digit mean accuracy for GPT-6 Luna. Limo grounds the same model in a composable spatial toolset and raises accuracy on the location queries to 99.3\%. Its advantage over competing MCP servers is largest where specialized routing, search, and join operators prevent the model from reimplementing spatial execution through fragile call sequences. These findings support a clear systems design: use the LLM for language interpretation and plan composition, use spatial systems for data and computation, and expose the boundary through compact, validated, single-responsibility tools. Limo is an initial step towards reliable location-aware LLM systems. We release the benchmark, toolset, and evaluation code to encourage the community to evaluate their own approaches, and we plan to maintain a public leaderboard to track progress. Future work will extend the benchmark across regions and evaluate queries at higher semantic and location complexity.

\section*{Statements and Declarations}

\noindent\textbf{Competing Interests.} The authors declare that they have no known competing financial interests or personal relationships that could have appeared to influence the work reported in this paper.

\noindent\textbf{Code and Data Availability.} The LimoBench queries, benchmark data, and evaluation code are available at \url{https://github.com/purduedb/limo-bench}.

\clearpage
\balance



\bibliographystyle{spmpsci}    
\bibliography{references}

\end{document}